\documentclass[10pt,twocolumn,letterpaper]{article}

\usepackage[pagenumbers]{cvpr}      

\usepackage{graphicx}
\usepackage{amsmath}
\usepackage{amssymb}
\usepackage{booktabs}
\usepackage{amsthm}
\newtheorem{theorem}{Theorem}[section]
\newtheorem{lemma}{Lemma}[section]
\usepackage{algorithm, algorithmic}
\usepackage{multirow}
\usepackage{makecell}

\usepackage[pagebackref,breaklinks,colorlinks]{hyperref}

\usepackage[capitalize]{cleveref}
\crefname{section}{Sec.}{Secs.}
\Crefname{section}{Section}{Sections}
\Crefname{table}{Table}{Tables}
\crefname{table}{Tab.}{Tabs.}

\def\cvprPaperID{10749} 
\def\confName{CVPR}
\def\confYear{2023}

\begin{document}

\title{Unsupervised Image Denoising with Score Function}

\author{Yutong Xie \\
Peking University 
\and
Mingze Yuan \\
Peking University 
\and
Bin Dong \\
Peking University 
\and 
Quanzheng Li \\
Massachusetts General Hospital and Harvard Medical School
}
\maketitle

\begin{abstract}
     Though achieving excellent performance in some cases, current unsupervised learning methods for single image denoising usually have constraints in applications. In this paper, we propose a new approach which is more general and applicable to complicated noise models. Utilizing the property of score function, the gradient of logarithmic probability, we define a solving system for denoising. Once the score function of noisy images has been estimated, the denoised result can be obtained through the solving system. Our approach can be applied to multiple noise models, such as the mixture of multiplicative and additive noise combined with structured correlation. Experimental results show that our method is comparable when the noise model is simple, and has good performance in complicated cases where other methods are not applicable or perform poorly.
\end{abstract}

\section{Introduction}
\label{sec:intro}

Image denoising \cite{dabov2006image, ramani2008monte, zhang2017beyond} has been studied for many years. Suppose $\boldsymbol{x}$ is a clean image, $\boldsymbol{y}$ is a noisy image of $\boldsymbol{x}$, and $p \left( \boldsymbol{y} \mid \boldsymbol{x} \right)$ is the noise model. Supervised learning methods try to train a model representing a mapping from $\boldsymbol{y}$ to $\boldsymbol{x}$. Due to the difficulty of collecting paired clean and noisy images in practice, methods in unsupervised learning manner are focus of research. Noise2Noise \cite{lehtinen2018noise2noise} is the first to use pairs of two different noisy images constructed from the same clean image to train a denoising model. Strictly speaking, Noise2Noise is not an unsupervised learning method. Collecting noisy pairs is still difficult. Despite all this, many other methods \cite{batson2019noise2self, krull2019noise2void, krull2020probabilistic, wu2020unpaired, xu2020noisy} are inspired from Noise2Noise or borrow the idea behind of it. These methods achieve good performance in some simple noise models.

The main drawback of current methods is the constraint of application. 
Once the noise model is complicated, they either are not applicable or have poor performance. Noisier2Noise \cite{moran2020noisier2noise} can only handle additive noise and pepper noise. Recorrupted-to-Recorrupted \cite{pang2021recorrupted} is limited to Gaussian noise. Neighbor2Neighbor \cite{huang2021neighbor2neighbor} requires that the noise model is pixel-wise independent and unbiased (\ie $\mathbb{E} \left[ \boldsymbol{y} \mid \boldsymbol{x} \right] = \boldsymbol{x}$). Noise2Score \cite{kim2021noise2score} applies Tweedie's Formula to image denoising and provides a unified framework for those noise models that follow exponential family distributions. However, the practical noise model will be more complicated. It may contain both multiplicative and additive noise, even combining with structural correlation.

\begin{figure*}
  \centering
    \includegraphics[width=0.75\linewidth]{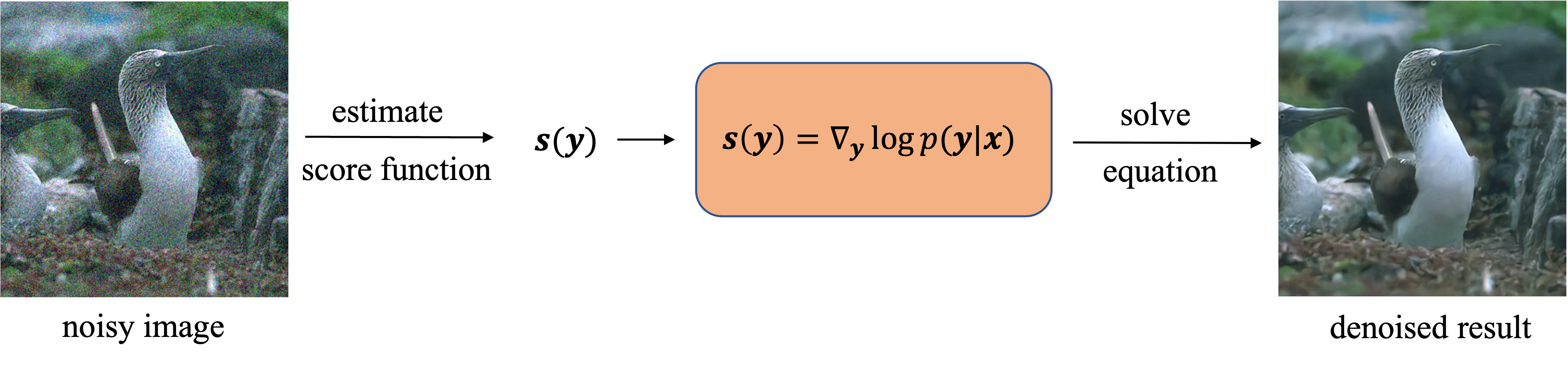}
    \caption{The overall flow of our approach. The first step is to estimate the score function $\boldsymbol{s}(\boldsymbol{y})$ and the second step is to solve an equation.}
    \label{fig:demo}
\end{figure*}

In this paper, we propose a new unified approach to handle more noise models. The key of our approach is a theoretical property about score function, $\nabla_{\boldsymbol{y}} \log p(\boldsymbol{y})$, which is shown in \cref{them:score-function}. This property indicates that the score function of $\boldsymbol{y}$ is the average of the score function of $\boldsymbol{y} \mid \boldsymbol{x}$ under the posterior distribution $p \left( \boldsymbol{x} \mid \boldsymbol{y}\right)$. Based on it, we define a system that the score function of $\boldsymbol{y}$ equals to the score function of $\boldsymbol{y} \mid \boldsymbol{x}$, which turns out to be an equation about $\boldsymbol{x}$ since $\boldsymbol{y}$ is known. We discover that the solution of this equation can be regarded as the denoised result of $\boldsymbol{y}$. Implementing our approach practically contains two steps: the first is to estimate the score function of $\boldsymbol{y}$ and the second is to solve the system defined above according to the specific noise model. The overall flow is shown in \cref{fig:demo}. All the model training is the estimation of score function, which is represented by a neural network. We adapt 
the amortized residual denoising autoencoder (AR-DAE) \cite{lim2020ar} to figure out it. More details can be seen in \cref{sec:estimation}.

Our approach is so powerful that as long as the score function of $\boldsymbol{y} \mid \boldsymbol{x}$ is known and the system is solvable, any kind of noise model is applicable. It means our approach is even able to solve sophisticated noise models such as mixture noise. Another advantage of our approach is that regardless of noise models, the training process of the score function neural network is identical. Therefore, once the assumption of the noise model does not hold or parameters of noise are corrected, we only change the equation system to be solved and resolve it without training the model again. In summary, our main contribution are: (1) We propose a general unsupervised approach for image denoising, which is based on the score function. (2) Experimental results show that our approach is competitive for simple noise models and achieves excellent performance for complicated noise models where other unsupervised methods is invalid.

\section{Related Works}
\label{sec:related_works}

Here we briefly review the existing deep learning methods for image denoising. When the pairs of clean images and noisy images are available, the supervised training \cite{zhang2017beyond} is to minimize $\mathbb{E}_{\boldsymbol{x}, \boldsymbol{y}} \left[ \mathrm{d} \left(\boldsymbol{x}, f(\boldsymbol{y}; \theta) \right) \right]$, where $f(\cdot; \theta)$ is a neural network and $\mathrm{d} \left( \cdot, \cdot\right)$ is a distance metric. Though supervised learning has great performance, the difficult acquisition for training data hampers its application in practice.

To avoid the issues of acquisition for paired clean and noisy images, unsupervised\footnote{Sometimes self-supervised is used} approaches are proposed
to use noisy image pairs to learn image denoising, which is started from Noise2Noise (N2N) \cite{lehtinen2018noise2noise}. While N2N can
achieve comparable results with supervised methods, collecting noisy image pairs from real world is still intractable. Motivated by N2N, the followed-up works try to learn image denoising with individual noisy images. Mask-based unsupervised approaches \cite{batson2019noise2self} design mask schemes for denoising on individual noisy images and then they train the network to predict the masked pixels according to noisy pixels in the input receptive field. Noise2Void \cite{krull2019noise2void} also proposes the blind-spot network (BSN) to avoid learning the identity function. Noisier2Noise \cite{moran2020noisier2noise}, Noisy-As-Clean \cite{xu2020noisy} and Recorrupted-to-Recorrupted \cite{pang2021recorrupted} require a single noisy realization of each training sample and a statistical model of the noise distribution. Noisier2Noise first generates a synthetic noisy sample from the statistical noise model, adds it to the already noisy image, and asks the network to predict the original noisy image from the doubly noisy image. Besides the blind-spot network design, Self2Self \cite{quan2020self2self} is proposed on blind denoising to generate paired data from a single noisy image by applying Bernoulli dropout. Recently, Neighbor2Neighbor \cite{huang2021neighbor2neighbor} proposes to create subsampled paired images based on the pixel-wise independent noise assumption and then a denoising network is trained on generated pairs, with additional regularization loss for better performance. Noise2Score \cite{kim2021noise2score} is another type of unsupervised methods that can deal with any noise model which follows an exponential family distribution. It utilize Tweedie's Formula \cite{robbins1956empirical, efron2011tweedie} and the estimation of score function to denoise noisy images.

\section{Method}
\label{sec:method}

In this section, we provide a more detailed description of the presented approach. The organization is as follows: In \cref{sec:theory} we introduce the basic theory and derive the specific algorithms for different noise models. We present the relationship between our method and Noise2Score in \ref{sec:relation}. Finally, we describe the method for estimating the score function $\nabla_{\boldsymbol{y}} \log p \left( \boldsymbol{y} \right)$ in \ref{sec:estimation}. All the full proofs of derivation are in Appendix.

\subsection{Basic Theory and Derivations}
\label{sec:theory}

\subsubsection{Basic Theory}

We begin with the following theorem.
\begin{theorem}
\label{them:score-function}
    Let $\boldsymbol{y} \sim \mathbf{y}$ and $\boldsymbol{x} \sim \mathbf{x}$ where $\mathbf{x}$ and $\mathbf{y}$ are two random variable, then the equation below holds:
    \begin{equation}
    \label{eq:score-function}
        \nabla_{\boldsymbol{y}} \log p\left(\boldsymbol{y}\right) = \int p\left(\boldsymbol{x} \mid \boldsymbol{y}\right) \nabla_{\boldsymbol{y}} \log p\left(\boldsymbol{y} \mid \boldsymbol{x}\right) \mathrm{d} \boldsymbol{x}
    \end{equation}
\end{theorem}
The proof of \cref{them:score-function} is in Appendix. In the task of image denoising, suppose $\boldsymbol{y} \in \mathbb{R}^{d}$ is the noisy image and $\boldsymbol{x} \in \mathbb{R}^{d}$ is the clean image where $d$ represents the number of image pixels. $p\left(\boldsymbol{x} \mid \boldsymbol{y}\right)$ represents the posterior distribution and $p\left(\boldsymbol{y} \mid \boldsymbol{x}\right)$ denotes the noise model. Our approach is inspired by \cref{eq:score-function}. For convenience, let $\boldsymbol{s}\left( \boldsymbol{y} \right) = \nabla_{\boldsymbol{y}} \log p\left(\boldsymbol{y}\right)$ and $\boldsymbol{f} \left(\boldsymbol{x}, \boldsymbol{y}\right) = \nabla_{\boldsymbol{y}} \log p\left(\boldsymbol{y} \mid \boldsymbol{x}\right)$, we define an equation about $\boldsymbol{x}$ as follows:
\begin{equation}
\label{eq:system}
    \boldsymbol{s}\left( \boldsymbol{y} \right) = \boldsymbol{f} \left(\boldsymbol{x}, \boldsymbol{y}\right).
\end{equation}
Assuming that $\boldsymbol{y}, \boldsymbol{s}\left( \boldsymbol{y} \right)$ and $\boldsymbol{f} \left(\boldsymbol{x}, \boldsymbol{y}\right)$ are known, then the solution of \cref{eq:system}, $\hat{\boldsymbol{x}}$, can be regarded as a denoised result of $\boldsymbol{y}$. Since $\boldsymbol{s}\left( \boldsymbol{y} \right)$ is the average of $\boldsymbol{f} \left(\boldsymbol{x}, \boldsymbol{y}\right)$ under the posterior distribution $p\left(\boldsymbol{x} \mid \boldsymbol{y}\right)$ according to \cref{eq:score-function}, $\hat{\boldsymbol{x}}$ should somehow be like the average of $\boldsymbol{x}$ conditioned on $\boldsymbol{y}$. Therefore, our approach is consist of two steps as shown in \cref{fig:demo}. The general algorithm is illustrated in \cref{alg:general}. 

\begin{algorithm}
\renewcommand{\algorithmicrequire}{\textbf{Input:}}
\renewcommand{\algorithmicensure}{\textbf{Output:}}
\caption{The general denoising process}
\label{alg:general}
\begin{algorithmic}[1]
    \REQUIRE noisy image $\boldsymbol{y}$ and $\boldsymbol{f} \left(\boldsymbol{x}, \boldsymbol{y}\right)$.
    \ENSURE $\hat{\boldsymbol{x}}$, the solution of \cref{eq:system}.
    \STATE Estimate $\boldsymbol{s}\left( \boldsymbol{y} \right)$.
    \STATE Solve \cref{eq:system}.
\end{algorithmic}  
\end{algorithm}

Next, we will derive the specific algorithm for solving $\boldsymbol{x}$ in \cref{eq:system} under different noise models, including additive Gaussian noise in \cref{sec:additive}, multiplicative noise in \cref{sec:multiplicative} and mixture noise in \cref{sec:mixture}. The estimation of score function $\boldsymbol{s}\left( \boldsymbol{y} \right)$ will be introduced in \cref{sec:estimation}.

\subsubsection{Additive Gaussian Noise}
\label{sec:additive}

Suppose the noise model is $\boldsymbol{y} = \boldsymbol{x} + \boldsymbol{\epsilon}$
where $\boldsymbol{\epsilon}$ follows a multi-variable Gaussian distribution with mean of $\boldsymbol{0}$ and covariance matrix of $\boldsymbol{\Sigma}$ denoted by $\mathcal{N}\left(\boldsymbol{0}, \boldsymbol{\Sigma}\right)$,
\ie 
\begin{equation}
    p \left( \boldsymbol{y} \mid \boldsymbol{x} \right) = \frac{\exp \left\{ - \frac{\left(\boldsymbol{y} - \boldsymbol{x} \right)^{\top} \boldsymbol{\Sigma}^{-1} \left(\boldsymbol{y} - \boldsymbol{x} \right)}{2} \right\}}{\sqrt{2 \pi}^{d}\left| \boldsymbol{\Sigma} \right|^{\frac{1}{2}}}  
\end{equation}
Then, we derive that
\begin{equation}
\begin{split}
    \boldsymbol{f} \left(\boldsymbol{x}, \boldsymbol{y}\right) = \nabla_{\boldsymbol{y}} \log p\left(\boldsymbol{y} \mid \boldsymbol{x}\right) = - \boldsymbol{\Sigma}^{-1} \left(\boldsymbol{y} - \boldsymbol{x} \right).
\end{split}
\end{equation}
We consider the following four kinds of $\boldsymbol{\Sigma}$:
\begin{enumerate}
    \item $\boldsymbol{\Sigma} = \sigma^2 \boldsymbol{I}$;
    \item $\boldsymbol{\Sigma} = \sigma^2 \boldsymbol{A}^{\top} \boldsymbol{A}$;
    \item $\boldsymbol{\Sigma} = \boldsymbol{\Sigma}\left(\boldsymbol{x} \right) = \mathrm{diag}\left({a \boldsymbol{x} + b\boldsymbol{1}}\right)^2$;
    \item $\boldsymbol{\Sigma} = \boldsymbol{\Sigma}\left(\boldsymbol{x} \right) = \boldsymbol{A}^{\top} \mathrm{diag}\left({a \boldsymbol{x} + b\boldsymbol{1}}\right)^2 \boldsymbol{A}$;
\end{enumerate}
where in the second and fourth cases, $\boldsymbol{A}$ usually represents a convolution transform which is used to describe the correlation between adjacent pixels. 

For the first and second cases, $\boldsymbol{\Sigma}$ is a constant matrix. By solving \cref{eq:system}, we have
\begin{equation}
\label{eq:gaussian-simple}
    \hat{\boldsymbol{x}} = \boldsymbol{\Sigma} \boldsymbol{s}\left( \boldsymbol{y} \right) + \boldsymbol{y}.
\end{equation}
While in the third and fourth cases, $\boldsymbol{\Sigma}$ is related to $\boldsymbol{x}$ and \cref{eq:gaussian-simple} is a fixed point equation. 
Therefore, we use an iterative trick to solve it as shown in \cref{alg:gaussian-fix}.
\begin{algorithm}
\renewcommand{\algorithmicrequire}{\textbf{Input:}}
\renewcommand{\algorithmicensure}{\textbf{Output:}}
\caption{An iterative trick to solve ${\boldsymbol{x}} = \boldsymbol{\Sigma}\left({\boldsymbol{x}} \right) \boldsymbol{s}\left( \boldsymbol{y} \right) + \boldsymbol{y}$}
\label{alg:gaussian-fix}
\begin{algorithmic}[1]
    \REQUIRE noisy image $\boldsymbol{y}$, $\boldsymbol{s}\left( \boldsymbol{y} \right)$, the parameters of $\boldsymbol{\Sigma} \left(\cdot\right)$ and the number of iterations $n$.
    \ENSURE $\hat{\boldsymbol{x}}$, the solution.
    \STATE Initial value of $\hat{\boldsymbol{x}}$ is set as $\boldsymbol{y}$.
    \FOR{$i = 1, ..., n$}
        \STATE $\hat{\boldsymbol{x}} \leftarrow \boldsymbol{\Sigma}\left(\hat{\boldsymbol{x}} \right) \boldsymbol{s}\left( \boldsymbol{y} \right) + \boldsymbol{y}$.
    \ENDFOR
\end{algorithmic}  
\end{algorithm}

\subsubsection{Multiplicative Noise}
\label{sec:multiplicative}
Firstly, we discuss three types of multiplicative noise model, Gamma, Poisson and Rayleigh noise. Then, we consider the situation where the convolution transform $\boldsymbol{A}$ exists.

\paragraph{Gamma Noise} Gamma Noise is constructed from Gamma distribution: $p(x; \alpha, \beta) = \frac{\beta^{\alpha}}{\Gamma\left( \alpha \right)} x^{\alpha - 1} e^{- \beta x}$.
We denote the Gamma distribution with parameters $\alpha$ and $\beta$ as $\mathcal{G}\left(\alpha, \beta \right)$. The noise model is defined as follows:
\begin{equation}
    \boldsymbol{y} = \boldsymbol{\eta} \odot \boldsymbol{x}, \quad {\eta}_{i} \sim  \mathcal{G}\left(\alpha, \alpha \right), \alpha > 1,
\end{equation}
where $\odot$ means component-wise multiplication, \ie
\begin{equation}
\label{eq:gamma-prob}
    p \left(\boldsymbol{y} \mid \boldsymbol{x} \right) = \prod_{i=1}^{d} \frac{\alpha^\alpha}{\Gamma\left( \alpha \right)} \left(\frac{y_i}{x_i}\right)^{\alpha - 1} \exp \left\{- \frac{\alpha y_i}{x_i}\right\} \cdot \frac{1}{x_i}.
\end{equation}
Then, we derive that
\begin{equation}
\label{eq:gamma-log}
\begin{split}
    \boldsymbol{f} \left(\boldsymbol{x}, \boldsymbol{y}\right) = \nabla_{\boldsymbol{y}} \log p\left(\boldsymbol{y} \mid \boldsymbol{x}\right) = \frac{\alpha - 1}{\boldsymbol{y}} - \frac{\alpha}{\boldsymbol{x}}.
\end{split}
\end{equation}
Here, the division is component-wise division. In the residual part of this paper, we neglect such annotation if it is not ambiguous. By solving \cref{eq:system}, we have
\begin{equation}
\label{eq:gamma-solution}
    \hat{\boldsymbol{x}} = \frac{\alpha \boldsymbol{y}}{\alpha - 1 - \boldsymbol{y} \odot \boldsymbol{s}\left( \boldsymbol{y}\right)}.
\end{equation}

\paragraph{Poisson Noise} Poisson Noise is constructed from Poisson distribution: $\mathrm{Pr}(x=k)=\frac{\lambda^{k}}{k !} e^{- \lambda}, k = 0, 1, \cdots$. 
We denote the Poisson distribution with parameters $\lambda$ as $\mathcal{P}\left(\alpha, \beta \right)$. The noise model is defined as follows:
\begin{equation}
    \boldsymbol{y} = \frac{1}{\lambda} \boldsymbol{\eta}, \quad {\eta}_{i} \sim  \mathcal{P}\left(\lambda x_{i} \right), \lambda > 0.
\end{equation}
\ie
\begin{equation}
\label{eq:poisson-prob}
    \mathrm{Pr} \left(\boldsymbol{y} \mid \boldsymbol{x} \right) = \prod_{i=1}^{d} \frac{\left( \lambda x_{i} \right)^{\lambda_{i} y_{i}}}{\left( \lambda y_{j} \right) !} e^{- \lambda x_{i}}.
\end{equation}
Then, we derive that
\begin{equation}
\label{eq:poisson-log}
\begin{split}
    \boldsymbol{f} \left(\boldsymbol{x}, \boldsymbol{y}\right) &= \nabla_{\boldsymbol{y}} \log \mathrm{Pr} \left(\boldsymbol{y} \mid \boldsymbol{x}\right) \\
    &= \lambda \log \left( \lambda \boldsymbol{x} \right) - \lambda \log \left( \lambda \boldsymbol{y} + \frac{1}{2} \right).
\end{split}
\end{equation}
By solving \cref{eq:system}, we have
\begin{equation}
\label{eq:poisson-solution}
    \hat{\boldsymbol{x}} = \left( \boldsymbol{y} + \frac{1}{2\lambda} \right) \odot \exp \left\{\frac{\boldsymbol{s}\left( \boldsymbol{y}\right)}{\lambda } \right\},
\end{equation}

\paragraph{Rayleigh Noise} Rayleigh Noise is constructed from Rayleigh distribution: $p(x; \sigma)= \frac{x}{\sigma^2} \exp\left\{-\frac{x^2}{2 \sigma^2}\right\}.$
We denote the Rayleigh distribution with parameters $\sigma$ as $\mathcal{R}\left(\sigma \right)$. The noise model is defined as follows:
\begin{equation}
    \boldsymbol{y} = \left( \boldsymbol{\eta} + \boldsymbol{1} \right) \odot \boldsymbol{x}, \quad {\eta}_{i} \sim  \mathcal{R}\left(\sigma \right), \sigma > 0.
\end{equation}
\ie
\begin{equation}
\label{eq:rayleigh-prob}
    p \left(\boldsymbol{y} \mid \boldsymbol{x} \right) = \prod_{i=1}^{d} \frac{1}{x_{i}} \frac{y_{i} - x_{i}}{x_{i} \sigma^2} \exp \left\{ - \frac{(y_i - x_i)^2}{2 x_i^2 \sigma^2}\right\}.
\end{equation}
Then, we derive that
\begin{equation}
\label{eq:rayleigh-log}
\begin{split}
    \boldsymbol{f} \left(\boldsymbol{x}, \boldsymbol{y}\right) &= \nabla_{\boldsymbol{y}} \log p \left(\boldsymbol{y} \mid \boldsymbol{x}\right) = \frac{1}{\boldsymbol{y} - \boldsymbol{x}} - \frac{\boldsymbol{y} - \boldsymbol{x}}{\sigma^2 \boldsymbol{x}^2}.
\end{split}
\end{equation}
Solving \cref{eq:system} directly is not easy. Here we provide an iterative algorithm to solve it. It is illustrated in \cref{alg:rayleigh-iterative}.
\begin{algorithm}
\renewcommand{\algorithmicrequire}{\textbf{Input:}}
\renewcommand{\algorithmicensure}{\textbf{Output:}}
\caption{An iterative method to solve \cref{eq:system} in the case of Rayleigh noise}
\label{alg:rayleigh-iterative}
\begin{algorithmic}[1]
    \REQUIRE noisy image $\boldsymbol{y}$, $\boldsymbol{s}\left( \boldsymbol{y} \right)$, the parameter of Rayleigh noise $\sigma$ and the number of iterations $n$.
    \ENSURE $\hat{\boldsymbol{x}}$, the solution of \cref{eq:system}.
    \STATE Initial value of $\hat{\boldsymbol{x}}$ is set as $\boldsymbol{y}$.
    \FOR{$i = 1, ..., n$}
        \STATE Compute $\boldsymbol{b} = \sigma^2 \boldsymbol{s}\left( \boldsymbol{y} \right) \odot \hat{\boldsymbol{x}}$
        \STATE Compute $\boldsymbol{t} = \frac{1}{2} \left( - \boldsymbol{b} + \sqrt{\boldsymbol{b} \odot \boldsymbol{b}  + 4 \sigma^2 \boldsymbol{1}}\right)$
        \STATE $\hat{\boldsymbol{x}} \leftarrow \frac{\boldsymbol{y}}{\boldsymbol{t} + \boldsymbol{1}}$.
    \ENDFOR
\end{algorithmic}  
\end{algorithm}

Now, we consider the situation where the convolution transform $\boldsymbol{A}$ exists. Suppose the noise model is represented by
\begin{equation}
    \boldsymbol{y} = \boldsymbol{A} \boldsymbol{z}, \quad \boldsymbol{z} =  N \left(\boldsymbol{x} \right),
\end{equation}
where $N\left(\boldsymbol{x} \right)$ can be any multiplicative noise model discussed above. 
Then, we have
\begin{equation}
\label{eq:multi-conv}
\begin{split}
    \nabla_{\boldsymbol{y}} \log p_{\mathbf{y}} \left(\boldsymbol{y} \mid \boldsymbol{x}\right) = \boldsymbol{A}^{-1, \top} \nabla_{\boldsymbol{z}} \log p_{\mathbf{z}} \left( \boldsymbol{A}^{-1} \boldsymbol{y} \mid \boldsymbol{x} \right).
\end{split}    
\end{equation}
To avoid confusion, we use subscripts to distinguish different distribution. Therefore, we can apply \cref{alg:multiplicative-correlation} to solve \cref{eq:system}, which is shown as follows.
\begin{algorithm}
\renewcommand{\algorithmicrequire}{\textbf{Input:}}
\renewcommand{\algorithmicensure}{\textbf{Output:}}
\caption{The general framework to solve \cref{eq:system} when the multiplicative noise model is combined with correlation.}
\label{alg:multiplicative-correlation}
\begin{algorithmic}[1]
    \REQUIRE $\boldsymbol{y}$, $\boldsymbol{s}\left( \boldsymbol{y} \right)$, $\boldsymbol{A}$ and $\tilde{\boldsymbol{f}}\left(\boldsymbol{x}, \boldsymbol{z} \right) = \nabla_{\boldsymbol{z}} \log p_{\mathbf{z}} \left( \boldsymbol{z} \mid \boldsymbol{x} \right)$.
    \ENSURE $\hat{\boldsymbol{x}}$, the solution of \cref{eq:system}.
    \STATE Computing $\tilde{\boldsymbol{s}} = \boldsymbol{A}^{\top} \boldsymbol{s} \left(\boldsymbol{y} \right)$.
    \STATE Computing $\boldsymbol{z} = \boldsymbol{A}^{-1} \boldsymbol{y}$.
    \STATE Solve $\tilde{\boldsymbol{s}} = \tilde{\boldsymbol{f}}\left(\boldsymbol{x}, \boldsymbol{z} \right)$ by the corresponding algorithm.
\end{algorithmic}  
\end{algorithm}

\subsubsection{Mixture Noise}
\label{sec:mixture}

In this paper, the mixture noise model is composed of a multiplicative noise and an additive Gaussian noise. It has the following form:
\begin{equation}
\label{eq:mixture-noise}
    \boldsymbol{y} = \boldsymbol{z} + \boldsymbol{\epsilon}, \epsilon \sim  \mathcal{N}\left(0, \sigma^2 \boldsymbol{I} \right)
\end{equation}
where $\boldsymbol{z} =  \boldsymbol{A}N\left(\boldsymbol{x} \right)$ is any multiplicative noise model that can be solved by our approach and $\boldsymbol{A}$ is either a convolution transform or identity matrix. It is easy to derive that
\begin{equation}
    p_{\mathbf{y}} \left( \boldsymbol{y} \mid \boldsymbol{x} \right) = \int p_{\mathbf{y}} \left( \boldsymbol{y} \mid \boldsymbol{z} \right) p_{\mathbf{z}} \left( \boldsymbol{z} \mid \boldsymbol{x} \right) \mathrm{d} \boldsymbol{z}.
\end{equation}
Generally speaking, $p_{\mathbf{y}} \left( \boldsymbol{y} \mid \boldsymbol{x} \right)$ has not an explicit analytical form. In this paper, we assume that the additive Gaussian noise is far smaller than the multiplicative noise. Thus, we utilize Taylor expansion to approximate $p_{\mathbf{y}} \left( \boldsymbol{y} \mid \boldsymbol{x} \right)$. We have the following conclusion:
\begin{equation}
\label{eq:mixture-taylor}
\begin{split}
    p_{\mathbf{y}} \left( \boldsymbol{y} \mid \boldsymbol{x} \right)
    \approx  p_{\mathbf{z}} \left( \bar{\boldsymbol{z}} \mid \mathbf{x} \right) + \nabla_{\boldsymbol{z}} p_{\mathbf{z}} \left( \bar{\boldsymbol{z}} \mid \boldsymbol{x} \right)^T \left( \boldsymbol{y} - \bar{\boldsymbol{z}} \right),
\end{split}
\end{equation}
where $\bar{\boldsymbol{z}} = \mathbb{E}\left[ \boldsymbol{z} \mid \boldsymbol{y} \right]$. Then, we can further derive that
\begin{equation}
\label{eq:mixture-score}
    \nabla_{\boldsymbol{y}} \log p_{\mathbf{y}} \left( \boldsymbol{y} \mid \boldsymbol{x} \right) \approx  \nabla_{\boldsymbol{z}} \log p_{\mathbf{z}} \left( \bar{\boldsymbol{z}} \mid \boldsymbol{x} \right).
\end{equation}
The full and rigorous derivations of \cref{eq:mixture-taylor} and \cref{eq:mixture-score} are in Appendix. Applying \cref{eq:gaussian-simple} in \cref{sec:additive}, we have $\bar{\boldsymbol{z}} = \boldsymbol{y} + \sigma^2 \boldsymbol{s}\left( \boldsymbol{y} \right)$. Thus, the equation to be solve is 
\begin{equation}
\label{eq:mixture-system}
   \boldsymbol{s}\left( \boldsymbol{y} \right) = \boldsymbol{f}\left(\boldsymbol{x}, \bar{\boldsymbol{z}} \right).
\end{equation}
The full denoising process is illustrated in \cref{alg:mixture}.
\begin{algorithm}
\renewcommand{\algorithmicrequire}{\textbf{Input:}}
\renewcommand{\algorithmicensure}{\textbf{Output:}}
\caption{The full denoising process for mixture noise in \cref{eq:mixture-noise}}
\label{alg:mixture}
\begin{algorithmic}[1]
    \REQUIRE $\boldsymbol{y}$, $\boldsymbol{s}\left( \boldsymbol{y} \right)$.
    \ENSURE $\hat{\boldsymbol{x}}$, the solution of \cref{eq:mixture-system}.
    \STATE Computing $\bar{\boldsymbol{z}} = \boldsymbol{y} + \sigma^2 \boldsymbol{s}\left( \boldsymbol{y} \right)$.
    \STATE Solve \cref{eq:mixture-system} through the corresponding algorithm discussed in \cref{sec:multiplicative}.
\end{algorithmic}  
\end{algorithm}

\subsection{Relation to Noise2Score}
\label{sec:relation}

Though our approach is derived directly from \cref{them:score-function}, it turns out to be connected closely with Noise2Score, which is also related to the score function. Specifically, when the noise model belongs to exponential family distributions, we can derive the same result as Tweedie's Formula from \cref{them:score-function}.

Suppose the noise model follows some exponential family distribution and has the following form:
\begin{equation}
\label{eq:exp-distribution}
    p \left( \boldsymbol{y} \mid \boldsymbol{x} \right) = b \left( \boldsymbol{y} \right) \exp \left\{ H(\boldsymbol{x})^{\top} T\left( \boldsymbol{y} \right) - a \left( \boldsymbol{x} \right) \right\},
\end{equation}
where $T\left( \boldsymbol{y} \right)$ and $(\boldsymbol{x})$ have the same dimensions, and $a \left( \boldsymbol{x} \right)$ and $b \left( \boldsymbol{y} \right)$ are both scalar functions. One of the properties of an exponential family distribution is that:
\begin{equation}
\label{eq:exp_property}
    H^{\prime}(\boldsymbol{x})^{\top} \mathbb{E} \left[ T(\boldsymbol{y}) \mid \boldsymbol{x} \right]  = a^{\prime} (\boldsymbol{x}).
\end{equation}
According to Bayesian Formula, we have the following derivation:
\begin{equation}
\begin{split}
    & p \left(\boldsymbol{x} \mid \boldsymbol{y} \right) = \frac{p \left( \boldsymbol{x} \right) p \left( \boldsymbol{y} \mid \boldsymbol{x} \right) }{p \left( \boldsymbol{y} \right) } \\
    =& p \left( \boldsymbol{x} \right) e^{- a \left( \boldsymbol{x} \right)} \exp \left( T\left( \boldsymbol{y} \right)^{\top} H(\boldsymbol{x}) + \log \frac{b \left( \boldsymbol{y} \right)}{p \left( \boldsymbol{y} \right)} \right).
\end{split}
\end{equation}
Therefore, $p \left(\boldsymbol{x} \mid \boldsymbol{y} \right)$ is also an exponential family distribution. According to \cref{eq:exp_property}, we can derive Tweedie's Formula that
\begin{align}
    & T^{\prime}\left( \boldsymbol{y} \right)^{\top} \mathbb{E} \left[ H(\boldsymbol{x}) \mid {\boldsymbol{y}}\right] = \nabla_{{\boldsymbol{y}}} \log p \left( {\boldsymbol{y}}\right) -  \nabla_{{\boldsymbol{y}}} \log b \left({\boldsymbol{y}}\right) \notag \\
    \Leftrightarrow & \boldsymbol{s} \left( \boldsymbol{y} \right) = \nabla_{{\boldsymbol{y}}} \log b \left({\boldsymbol{y}}\right) + T^{\prime}\left( \boldsymbol{y} \right)^{\top} \mathbb{E} \left[ H(\boldsymbol{x}) \mid {\boldsymbol{y}}\right].
\end{align}
Applying \cref{eq:exp-distribution} to \cref{eq:score-function}, we also have
\begin{equation}
\label{eq:tweedie-score}
\begin{split}
\boldsymbol{s} \left( \boldsymbol{y} \right) = \nabla_{{\boldsymbol{y}}} \log b \left({\boldsymbol{y}}\right) + T^{\prime}\left( \boldsymbol{y} \right)^{\top} \mathbb{E} \left[ H(\boldsymbol{x}) \mid {\boldsymbol{y}}\right].
\end{split}
\end{equation}
Therefore, we prove Tweedie's Formula directly from \cref{them:score-function}.
Solving \cref{eq:system} will derive the same result as Noise2Score when the noise model follows an exponential family distribution. The specific conclusions of Gaussian, Gamma and Poisson noise can be seen in Appendix.

From the analysis above, we can come to a conclusion that our approach covers Noise2Score. However, based on Tweedie's Formula, Noise2Score cannot address noise models beyond exponential family distributions. Instead, our approach derived directly from \cref{them:score-function} can tackle more complicated cases such like Rayleigh noise and other examples discussed in \ref{sec:theory}.

\subsection{Estimation of Score Function}
\label{sec:estimation}

So far, we assume that the score function of $\boldsymbol{y}$, $\boldsymbol{s}\left( \boldsymbol{y} \right)$, is known. However, it is usually unknown and should be estimated from the dataset of noisy images $\left\{ \boldsymbol{y} \right\}$. We use the same method, the amortized residual Denoising Auto Encoder (AR-DAE) \cite{lim2020ar} discussed in Noise2Score. Suppose $\boldsymbol{s}\left(\cdot; \theta\right)$ is a neural network used to represent the score function of $\boldsymbol{y}$. The following objective function is used to train the model:
\begin{equation}
    L = \mathbb{E}_{\boldsymbol{y}, \boldsymbol{u} \sim \mathcal{N}\left(\boldsymbol{0}, \boldsymbol{I}\right)} \left\| \boldsymbol{u} + \sigma_a \boldsymbol{s}\left(\boldsymbol{y} + \sigma_a \boldsymbol{u}; \theta\right) \right\|_2^2,
\end{equation}
where $\sigma_a$ is a fixed value. Given $\sigma_a$, the optimal model $\boldsymbol{s}(\boldsymbol{y}; \theta^*)$ that minimizes $L$ is the score function of perturbed $\boldsymbol{y}$, $\boldsymbol{y} + \sigma_a \boldsymbol{u}$. In other words, we can approximate the score function $\boldsymbol{s}\left( \boldsymbol{y} \right)$ by using a sufficiently small $\sigma_a$. Related analysis can also be seen in \cite{hyvarinen2005estimation, vincent2011connection, alain2014regularized}. During the training process, the value of $\sigma_a$ will be decreasing gradually to a very small value. This progressive process is helpful to numerically stabilize the model training. The only model training is to estimate $\boldsymbol{s}\left( \boldsymbol{y} \right)$ by $\boldsymbol{s}\left(\cdot; \theta\right)$, which is served as the first step of our approach. After the score function model is trained, we apply the denoising algorithms given in \cref{sec:theory} to obtain denoised results and no more training is required.

\begin{table*}
  \centering
  \begin{tabular}{@{}llccccc@{}}
    \toprule
    No. & Noise Model & SL & Nr2N & Nb2Nb & N2S & Ours \\
    \midrule
    1 & $\boldsymbol{y} = \boldsymbol{x} + \boldsymbol{\epsilon}, \boldsymbol{\epsilon} \sim \mathcal{N} \left( \boldsymbol{0}, \sigma^2 \boldsymbol{I} \right)$ & $\checkmark$ & $\checkmark$ & $\checkmark$ & $\checkmark$ & $\checkmark$ \\
    
    2 & $\boldsymbol{y} = \boldsymbol{x} + \boldsymbol{\epsilon}, \boldsymbol{\epsilon} \sim \mathcal{N} \left( \boldsymbol{0}, \sigma^2 \boldsymbol{A}^{\top} \boldsymbol{A} \right)$ & $\checkmark$ & $\checkmark$ & $\times^{*}$ & $\checkmark$ & $\checkmark$ \\
    
    3 & $\boldsymbol{y} = \boldsymbol{x} + \boldsymbol{\epsilon}, \boldsymbol{\epsilon} \sim \mathcal{N} \left( \boldsymbol{0}, \mathrm{diag}\left({a \boldsymbol{x} + b\boldsymbol{1}}\right)^2 \right)$ & $\checkmark$ & $\checkmark$ & $\checkmark$ & $\times$ & $\checkmark$ \\
    
    4 & $\boldsymbol{y} = \boldsymbol{x} + \boldsymbol{\epsilon}, \boldsymbol{\epsilon} \sim \mathcal{N} \left( \boldsymbol{0}, \boldsymbol{A}^{\top} \mathrm{diag}\left({a \boldsymbol{x} + b\boldsymbol{1}}\right)^2 \boldsymbol{A} \right)$ & $\checkmark$ & $\checkmark$ & $\times^{*}$ & $\times$ & $\checkmark$ \\

    
    5 & $\boldsymbol{y} = \boldsymbol{\eta} \odot \boldsymbol{x}, {\eta}_{i} \sim  \mathcal{G}\left(\alpha, \alpha \right)$ & $\checkmark$ & $\times$ & $\checkmark$ & $\checkmark$ & $\checkmark$ \\

    6 & $\boldsymbol{y} = \boldsymbol{A} \boldsymbol{\eta} \odot \boldsymbol{x}, {\eta}_{i} \sim  \mathcal{G}\left(\alpha, \alpha \right)$ & $\checkmark$ & $\times$ & $\checkmark^{*}$ & $\checkmark$ & $\checkmark$ \\

    7 & $\boldsymbol{y} = \frac{1}{\lambda} \boldsymbol{\eta}, {\eta}_{i} \sim  \mathcal{P}\left(\lambda x_{i} \right)$ & $\checkmark$ & $\times$ & $\checkmark$ & $\checkmark$ & $\checkmark$ \\

    8 & $\boldsymbol{y} = \frac{1}{\lambda} \boldsymbol{A}\boldsymbol{\eta}, {\eta}_{i} \sim  \mathcal{P}\left(\lambda x_{i} \right)$ & $\checkmark$ & $\times$ & $\checkmark^*$ & $\checkmark$ & $\checkmark$ \\

    9 & $\boldsymbol{y} = \left( \boldsymbol{\eta} + \boldsymbol{1} \right) \odot \boldsymbol{x}, {\eta}_{i} \sim  \mathcal{R}\left(\sigma \right)$ & $\checkmark$ & $\times$ & $\times^*$ & $\times$ & $\checkmark$ \\

    10 & $\boldsymbol{y} =\boldsymbol{A} \left( \boldsymbol{\eta} + \boldsymbol{1} \right) \odot \boldsymbol{x}, {\eta}_{i} \sim  \mathcal{R}\left(\sigma \right)$ & $\checkmark$ & $\times$ & $\times^*$ & $\times$ & $\checkmark$ \\

    
    11 & $\boldsymbol{y} =  \boldsymbol{\eta} \odot \boldsymbol{x} + \boldsymbol{\epsilon}, {\eta}_{i} \sim  \mathcal{G}\left(\alpha, \alpha \right), \epsilon \sim  \mathcal{N}\left(0, \sigma^2 \boldsymbol{I} \right)$ & $\checkmark$ & $\times$ & $\checkmark$ & $\times$ & $\checkmark$ \\

    12 & $\boldsymbol{y} =  \boldsymbol{A} \boldsymbol{\eta} \odot \boldsymbol{x} + \boldsymbol{\epsilon}, {\eta}_{i} \sim  \mathcal{G}\left(\alpha, \alpha \right), \epsilon \sim  \mathcal{N}\left(0, \sigma^2 \boldsymbol{I} \right)$ & $\checkmark$ & $\times$ & $\times^*$ & $\times$ & $\checkmark$ \\

    13 & $\boldsymbol{y} =  \frac{1}{\lambda}\boldsymbol{\eta} + \boldsymbol{\epsilon}, {\eta}_{i} \sim  \mathcal{P}\left(\lambda x_i \right), \epsilon \sim  \mathcal{N}\left(0, \sigma^2 \boldsymbol{I} \right)$ & $\checkmark$ & $\times$ & $\checkmark$ & $\times$ & $\checkmark$ \\

    14 & $\boldsymbol{y} =  \frac{1}{\lambda} \boldsymbol{A} \boldsymbol{\eta} + \boldsymbol{\epsilon}, {\eta}_{i} \sim  \mathcal{P}\left(\lambda x_i \right), \epsilon \sim  \mathcal{N}\left(0, \sigma^2 \boldsymbol{I} \right)$ & $\checkmark$ & $\times$ & $\times^*$ & $\times$ & $\checkmark$ \\

    15 & $\boldsymbol{y} = \left( \boldsymbol{\eta} + \boldsymbol{1} \right) \odot \boldsymbol{x}, {\eta}_{i} \sim  \mathcal{R}\left(\sigma \right), \epsilon \sim  \mathcal{N}\left(0, \sigma^2 \boldsymbol{I} \right)$ & $\checkmark$ & $\times$ & $\times^*$ & $\times$ & $\checkmark$ \\

    16 & $\boldsymbol{y} =  \boldsymbol{A} \left( \boldsymbol{\eta} + \boldsymbol{1} \right) \odot \boldsymbol{x}, {\eta}_{i} \sim  \mathcal{R}\left(\sigma \right), \epsilon \sim  \mathcal{N}\left(0, \sigma^2 \boldsymbol{I} \right)$ & $\checkmark$ & $\times$ & $\times^*$ & $\times$ & $\checkmark$ \\
    
    \bottomrule
  \end{tabular}
  \caption{The application range of different methods including supervised learning (SL), Noisier2Noise (Nr2N), Neighbor2Neighbor (Nb2Nb), Noise2Score (N2S) and ours. $\checkmark$ means applicable and $\times$ means not applicable or incapable to perform. For Neighbor2Neighbor, $\checkmark^*$ means that direct application is not feasible but indirect application is; $\times^*$ means that the application is not feasible but model training is executable.}
  \label{tab:application}
\end{table*}

\section{Experiment}
\label{sec:experiment}

We conduct extensive experiments to evaluate our approach, including additive Gaussian noise, multiplicative noise and mixture noise. 

\paragraph{Dataset and Implementation Details} We evaluate the proposed method for color images in the three benchmark datasets containing RGB natural images: Kodak dataset, CBSD68 \cite{martin2001database} and CSet9. DIV2K \cite{timofte2018ntire} and CBSD500 dataset \cite{chaudhary2019comparative} are used as training datasets. The synthetic noise images for each noise model are generated and fixed through the training process. For the sake of fair comparison, we use the same modified U-Net \cite{dhariwal2021diffusion} for all methods. When training, we randomly clip the training images to patches with the resolution of $128 \times 128$. AdamW optimizer \cite{loshchilov2017decoupled} is used to train the network. We train each model for 5000 steps with the batch size of 32. To reduce memory, we utilize the tricks of cumulative gradient and mixed precision training. The learning rate is initialized to $1 \times 10^{-4}$ for first 4000 steps and it is decreased to $1 \times 10^{-5}$ for final 1000 steps. All the models are implemented in PyTorch \cite{paszke2017automatic} with NVidia V100. The pixel value range of all clean images is $[0, 255]$ and the parameters of noise models are built on it. Noisy images will be scaled when fed into the network. When an iterative algorithm is needed to solve \cref{eq:system}, we set the number of iterations as $10$. The more details of implementation are described in Appendix.

\paragraph{Baseline and Comparison Methods} We use supervised learning with MSE loss as the baseline model. Noisier2Noise and Neighbor2Neighbor are used as comparison methods. Since our approach is identical to Noise2Score when the noise model follows exponential family distributions, we do not compare to it through metrics. Because Noisier2Noise can only be applied to additive noise, we do not train models by Noisier2Noise for other noise models. Though Neighbor2Neighbor is not suitable for some noise models from the perspective of theoretical analysis, we still train corresponding models and report its results. \Cref{tab:application} shows the comparison of application range for different methods, based on which our experiments are conducted. Only supervised learning and our approach can handle all noise models listed in \cref{tab:application}.

\paragraph{Parameters of Noise Models} Here, we emphasize that for all noise models in our experiments, $\boldsymbol{A}$ is set as a $3 \times 3$ convolution transform with the kernel of 
\begin{equation}
    \left(
    \begin{matrix} 
    0.05 & 0.1 & 0.05 \\
    0.1 & 0.4 & 0.1 \\
    0.05 & 0.1 & 0.05 
    \end{matrix}
    \right)
\end{equation}
if it is used. The additive Gaussian noise in every mixture noise model is set as $\mathcal{N} \left(\boldsymbol{0}, 100 \boldsymbol{I} \right)$. Other parameters will be given later. Finally, all parameters are assumed to be known in our experiments.

\begin{figure*}
  \centering
  \includegraphics[width=0.85\linewidth]{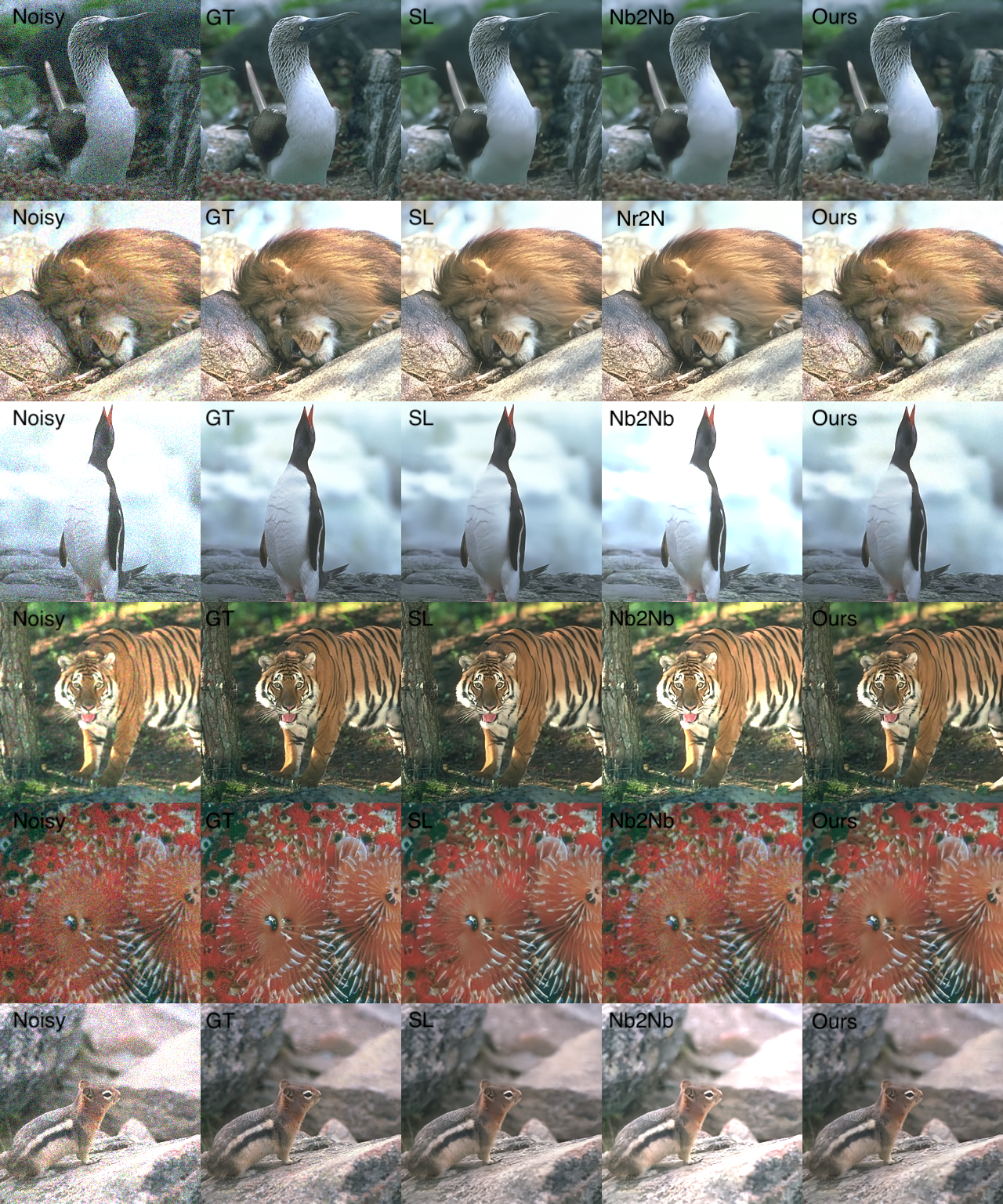}
  \caption{Qualitative Comparison using CBSD68 dataset (cropped to $256 \times 256$). From the first row to the last: (1) Gaussian noise, $\sigma=25$ w/o $\boldsymbol{A}$; (2) Gaussian noise, $\sigma=25$ w/ $\boldsymbol{A}$; (3) Rayleigh noise, $\sigma=0.3$ w/o $\boldsymbol{A}$; (4) Rayleigh noise, $\sigma=0.3$ w/ $\boldsymbol{A}$; (5) Poisson noise, $\lambda=0.2$ w/ $\boldsymbol{A}$ added by Gaussian Noise with $\sigma=10$; (6) Rayleigh noise, $\sigma=0.3$ w/o $\boldsymbol{A}$ added by Gaussian Noise with $\sigma=10$. Noisy: noisy image, GT: ground-truth image, SL: supervised learning, Nb2Nb: Neighbor2Neighbor, Nr2N: Noisier2Noise.}
  \label{fig:results}
\end{figure*}

\paragraph{Additive Gaussian Noise} Using additive Gaussian noise, we consider four kinds of noise models with different $\boldsymbol{\Sigma}$ corresponding from No.$1$ to No.$4$ in \cref{tab:application}. Our method is compared to supervised learning, Noisier2Noise and Neighbor2Neighbor as shown in \cref{tab:additive-gaussian-noise}. For the first two noise models $\sigma$ is $25$, and for the rest $a = 0.98$ and $b = 25$. As expected, supervised learning performs best. In the cases without $\boldsymbol{A}$ Neighbor2Neighbor is the best among three other unsupervised learning methods. However, in the cases with $\boldsymbol{A}$ Neighbor2Neighbor performs very poorly. Our approach outperform other unsupervised learning methods in the second noise model and is competitive in the whole. We show some results of the first two noise models in \cref{fig:results}.

\begin{table}
  \centering
  \begin{tabular}{@{}cccccc@{}}
    \toprule
    Params & Dataset & SL & Nr2N & Nb2Nb & Ours \\
    \midrule
    \multirow{3}*{\makecell{$\sigma = 25$ \\ w/o $\boldsymbol{A}$}} & Kodak & 32.44 & 31.80 &  31.96 & 31.92 \\
    ~ & CSet9 & 30.27 & 29.75 &  29.90 & 29.86 \\
    ~ & CBSD68 & 31.32 & 30.84 &  30.92 & 30.91 \\
    \midrule
    \multirow{3}*{\makecell{$\sigma = 25$ \\ w/ $\boldsymbol{A}$}} & Kodak & 34.80 & 33.87 &  27.89 & 33.99 \\
    ~ & CSet9 & 32.53 & 32.02 &  27.88 & 32.16 \\
    ~ & CBSD68 & 34.08 & 33.40 &  27.86 & 33.45 \\
    \midrule
    \multirow{3}*{\makecell{$a = 0.98$ \\ $b = 25$ \\ w/o $\boldsymbol{A}$}} & Kodak & 30.92 & 30.07 & 30.42 & 29.68 \\
    ~ & CSet9 & 28.69 & 28.02 &  28.22 & 27.58 \\
    ~ & CBSD68 & 29.68 & 28.99 &  29.29 & 28.70 \\
    \midrule
    \multirow{3}*{\makecell{$a = 0.98$ \\ $b = 25$ \\ w/ $\boldsymbol{A}$}} & Kodak & 33.02 & 32.28 &  25.02 & 31.88 \\
    ~ & CSet9 & 30.76 & 29.98 &  24.44 & 29.50 \\
    ~ & CBSD68 & 32.11 & 31.55 &  25.01 & 31.14 \\
    \bottomrule
  \end{tabular}
  \caption{Quantitative comparison for various parameters of $\boldsymbol{\Sigma}$ in additive Gaussian noise using different methods in terms of PNSR (dB).}
  \label{tab:additive-gaussian-noise}
\end{table}

\paragraph{Multiplicative Noise} We consider the combination of three various multiplicative noise model (Gamma, Poisson and Rayleigh) and a convolution transform $\boldsymbol{A}$. They are corresponding from No.$5$ to No.$10$ in \cref{tab:application}. Our method is compared to supervised learning and Neighbor2Neighbor and the results are shown in \cref{tab:multipicative-noise}. Because Noisier2Noise can not address such multiplicative noise models, we neglect it. Though the noise is not pixel-wise independent when the convolution transform exists, we can execute its inverse transform on $\boldsymbol{y}$ so that the requirement of pixel-wise independence is satisfied for Neighbor2Neighbor. Therefore, tackling noise models with a convolution transform is equivalent to the situations without $\boldsymbol{A}$. That is why we do not provide corresponding metrics result for Neighbor2Neighbor in \cref{tab:multipicative-noise}. We set $\alpha$ as $26$ for the Gamma noise, $\lambda$ as $0.2$ for the Poisson noise, and $\sigma$ as $0.3$ for the Rayleigh noise. When the noise model is based on Gamma or Poisson noise, it is unbiased, \ie $\mathbb{E} \left[ \boldsymbol{y} \mid \boldsymbol{x}\right] = \boldsymbol{x}$. In these cases, Neighbor2Neighbor is better than ours. However, when the noise model is based on Rayleigh noise which is biased our approach still has excellent performance while Neighbor2Neighbor is poor. The comparison in \cref{fig:results} also shows that our approach provides much better denoised results in the cases of Rayleigh noise.

\begin{table}
  \centering
  \begin{tabular}{@{}ccccc@{}}
    \toprule
    Params & Dataset & SL & Nb2Nb & Ours \\
    \midrule
    \multirow{3}*{\makecell{Gamma \\ $\alpha = 26$ \\ w/o $\boldsymbol{A}$}} 
      & Kodak & 33.51 & 32.98 &  32.61 \\
    ~ & CSet9 & 30.72 &  30.33 & 29.89 \\
    ~ & CBSD68 & 32.44 & 31.97 & 31.51 \\
    \midrule
    \multirow{3}*{\makecell{Gamma \\ $\alpha = 26$ \\ w/ $\boldsymbol{A}$}} 
      & Kodak & 33.02 & - &  31.90 \\
    ~ & CSet9 & 30.41 &  - & 29.26 \\
    ~ & CBSD68 & 32.15 & - & 30.63 \\
    \midrule
    \multirow{3}*{\makecell{Poisson \\ $\lambda = 0.2$ \\ w/o $\boldsymbol{A}$}} 
      & Kodak & 32.90 & 32.50 &  32.38 \\
    ~ & CSet9 & 30.56 &  30.17 & 29.98 \\
    ~ & CBSD68 & 31.87 & 31.48 & 31.31 \\
    \midrule
    \multirow{3}*{\makecell{Poisson \\ $\lambda = 0.2$ \\ w/ $\boldsymbol{A}$}}
      & Kodak & 32.55 & - &  31.84 \\
    ~ & CSet9 & 30.27 &  - & 29.48 \\
    ~ & CBSD68 & 31.64 & - & 30.56 \\
    \midrule
    \multirow{3}*{\makecell{Rayleigh \\ $\sigma = 0.3$ \\ w/o $\boldsymbol{A}$}} 
      & Kodak & 35.29 & 16.55 &  34.25 \\
    ~ & CSet9 & 32.44 &  15.16 & 31.45 \\
    ~ & CBSD68 & 34.39 & 16.74 & 33.34 \\    
    \midrule
    \multirow{3}*{\makecell{Rayleigh \\ $\sigma = 0.3$ \\ w/ $\boldsymbol{A}$}} 
      & Kodak & 34.63 & - &  32.87 \\
    ~ & CSet9 & 31.97 &  - & 30.30 \\
    ~ & CBSD68 & 33.94 & - & 31.85 \\
    \bottomrule
  \end{tabular}
  \caption{Quantitative comparison for various multiplicative noise models using different methods in terms of PNSR (dB). For Neighbor2Neighbor (Nb2Nb), if the noise model is constructed with $\boldsymbol{A}$, it can be regarded as the one without $\boldsymbol{A}$ through $\boldsymbol{A}^{-1} \boldsymbol{y}$. Thus we do not provide the metrics.}
  \label{tab:multipicative-noise}
\end{table}

\paragraph{Mixture Noise} We also consider the combination of three various multiplicative noise model (Gamma, Poisson and Rayleigh) and a convolution transform $\boldsymbol{A}$. For each one, additive Gaussian noise with $\boldsymbol{\Sigma} = 100 \boldsymbol{I}$ is added to construct mixture noise models. They are corresponding from No.$11$ to No.$16$ in \cref{tab:application}. Because of the same reason discussed before, our method is compared to supervised learning and Neighbor2Neighbor, and Noisier2Noise is neglected. The experimental results are shown in \cref{tab:mix-noise}. Due to the additive Gaussian Noise, Neighbor2Neighbor are not able to handle the cases with $\boldsymbol{A}$ through the inverse convolution transform. The correlation of noise hampers the performance of Neigh2bor2Neighbor. Except the first noise model in \cref{tab:mix-noise}, our approach all outperforms Neighbor2Neighbor and achieve excellent performance which is near supervised learning. The comparison in \cref{fig:results} also confirm that our approach provides good reconstruction results.

\begin{table}
  \centering
  \begin{tabular}{@{}ccccc@{}}
    \toprule
    Params & Dataset & SL & Nb2Nb & Ours \\
    \midrule
    \multirow{3}*{\makecell{Gamma \\ $\alpha = 26$ \\ w/o $\boldsymbol{A}$}} 
      & Kodak & 32.86 & 32.30 &  32.13 \\
    ~ & CSet9 & 30.23 &  29.80 & 29.61 \\
    ~ & CBSD68 & 31.70 & 31.21 & 31.08 \\
    \midrule
    \multirow{3}*{\makecell{Gamma \\ $\alpha = 26$ \\ w/ $\boldsymbol{A}$}} 
      & Kodak & 31.59 & 26.93 &  30.74 \\
    ~ & CSet9 & 29.30 &  25.06 & 28.40 \\
    ~ & CBSD68 & 30.42 & 26.33 & 29.58 \\
    \midrule
    \multirow{3}*{\makecell{Poisson \\ $\lambda = 0.2$ \\ w/o $\boldsymbol{A}$}} 
      & Kodak & 32.49 & 32.03 &  32.10 \\
    ~ & CSet9 & 30.18 &  29.74 & 29.70 \\
    ~ & CBSD68 & 31.40 & 30.98 & 31.03 \\
    \midrule
    \multirow{3}*{\makecell{Poisson \\ $\lambda = 0.2$ \\ w/ $\boldsymbol{A}$}} 
      & Kodak & 31.44 & 26.82 &  31.11 \\
    ~ & CSet9 & 29.36 &  25.48 & 28.80 \\
    ~ & CBSD68 & 30.32 & 26.32 & 29.85 \\
    \midrule
    \multirow{3}*{\makecell{Rayleigh \\ $\sigma = 0.3$ \\ w/o $\boldsymbol{A}$}} 
      & Kodak & 34.38 & 16.54 &  33.54 \\
    ~ & CSet9 & 31.79 &  15.15 & 30.94 \\
    ~ & CBSD68 & 33.39 & 16.71 & 32.68 \\    
    \midrule
    \multirow{3}*{\makecell{Rayleigh \\ $\sigma = 0.3$ \\ w/ $\boldsymbol{A}$}} 
      & Kodak & 32.83 & 16.40 &  31.14 \\
    ~ & CSet9 & 30.48 &  15.06 & 28.94 \\
    ~ & CBSD68 & 31.74 & 16.56 & 30.23 \\
    \bottomrule
  \end{tabular}
  \caption{Quantitative comparison for various mixture noise models using different methods in terms of PNSR (dB). For each noise model, additive Gaussian noise with $\boldsymbol{\Sigma}=100 \boldsymbol{I}$ is added.}
  \label{tab:mix-noise}
\end{table}

\paragraph{Robustness Evaluation}

\begin{figure}
  \centering
  \includegraphics[width=0.8\linewidth]{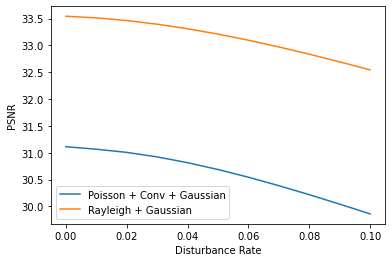}
  \caption{PSNR V.S. disturbance rate for No.14 and No.15 noise models}
  \label{fig:robust}
\end{figure}

To apply our approach, the parameters of noise models have to be known beforehand. Therefore their precision may impact on the performance. We choose No.14 and No.15 noise models in \cref{tab:application} as examples to show the robustness to parameters' precision. Suppose $k$ is one of parameters, we disturb it by $(a + 1)k$ where $a \sim \mathcal{N}(0, r^2)$. We call $r$ as the disturbance rate. The result of PSNR is shown in \cref{fig:robust}, which displays the robustness. Even $r = 0.1$, the value of PSNR only reduces about 1 dB.

\section{Conclusion}
\label{sec:conclusion}

In this paper, we propose a new approach for unsupervised image denoising. The key part is \cref{them:score-function}. Based on it, we construct an equation, \cref{eq:system} about the clean image $\boldsymbol{x}$ and the noisy image $\boldsymbol{y}$. After the score function of $\boldsymbol{y}$ is estimated, the denoised result can be obtained by solving the equation. Our approach can be applied to many different noise model as long as \cref{eq:system} is solvable. The denoising performance is competitive for simple noise models and excellent for complicated ones. We hope that this work is helpful to figure out sophisticated image denoising problems in practice.



\appendix

\section{Proofs}

\subsection{The proof of \cref{eq:score-function} in Sec. 3.1.1}

\begin{equation}
    \nabla_{\boldsymbol{y}} \log p\left(\boldsymbol{y}\right) = \int p\left(\boldsymbol{x} \mid \boldsymbol{y}\right) \nabla_{\boldsymbol{y}} \log p\left(\boldsymbol{y} \mid \boldsymbol{x}\right) \mathrm{d} \boldsymbol{x} \tag{1}
\end{equation}

\begin{proof}
Our derivation begins with the right part of \cref{eq:score-function}.
\begin{align*}
& \int p\left(\boldsymbol{x} \mid \boldsymbol{y}\right) \nabla_{\boldsymbol{y}} \log p\left(\boldsymbol{y} \mid \boldsymbol{x}\right) \mathrm{d} \boldsymbol{x} \\
= & \int p\left(\boldsymbol{x} \mid \boldsymbol{y}\right)  \nabla_{\boldsymbol{y}} \log \frac{ p\left(\boldsymbol{x} \mid \boldsymbol{y}\right)  p\left(\boldsymbol{y}\right)}{p\left(\boldsymbol{x}\right) }\mathrm{d} \boldsymbol{x} \\
= &  \int p\left(\boldsymbol{x} \mid \boldsymbol{y}\right) \left[ \nabla_{\boldsymbol{y}} \log p\left(\boldsymbol{x} \mid \boldsymbol{y}\right)  + \nabla_{\boldsymbol{y}} \log  p\left(\boldsymbol{y}\right) \right. \\ 
& \quad \quad \quad \left. - \nabla_{\boldsymbol{y}} \log  p\left(\boldsymbol{x}\right) \right] \mathrm{d} \boldsymbol{x} \\
= &  \int p\left(\boldsymbol{x} \mid \boldsymbol{y}\right) \left[ \nabla_{\boldsymbol{y}} \log p\left(\boldsymbol{x} \mid \boldsymbol{y}\right)  + \nabla_{\boldsymbol{y}} \log  p\left(\boldsymbol{y}\right) \right] \mathrm{d} \boldsymbol{x} \\ 
= &  \int p\left(\boldsymbol{x} \mid \boldsymbol{y}\right) \nabla_{\boldsymbol{y}} \log p\left(\boldsymbol{x} \mid \boldsymbol{y}\right) \mathrm{d} \boldsymbol{x}  \\
& \quad \quad \quad +  \nabla_{\boldsymbol{y}} \log  p\left(\boldsymbol{y}\right)  \int p\left(\boldsymbol{x} \mid \boldsymbol{y}\right) \mathrm{d} \boldsymbol{x} \\
= &  \int p\left(\boldsymbol{x} \mid \boldsymbol{y}\right) \nabla_{\boldsymbol{y}} \log p\left(\boldsymbol{x} \mid \boldsymbol{y}\right) \mathrm{d} \boldsymbol{x} + \nabla_{\boldsymbol{y}} \log  p\left(\boldsymbol{y}\right) \\
\end{align*}
Now, we prove that $\int p\left(\boldsymbol{x} \mid \boldsymbol{y}\right) \nabla_{\boldsymbol{y}} \log p\left(\boldsymbol{x} \mid \boldsymbol{y}\right) \mathrm{d} \boldsymbol{x} = 0$:
\begin{align*}
& \int p\left(\boldsymbol{x} \mid \boldsymbol{y}\right) \nabla_{\boldsymbol{y}} \log p\left(\boldsymbol{x} \mid \boldsymbol{y}\right) \mathrm{d} \boldsymbol{x} \\
= & \int p\left(\boldsymbol{x} \mid \boldsymbol{y}\right) \frac{1}{ p\left(\boldsymbol{x} \mid \boldsymbol{y}\right)} \nabla_{\boldsymbol{y}} p\left(\boldsymbol{x} \mid \boldsymbol{y}\right) \mathrm{d} \boldsymbol{x} \\
= & \int \nabla_{\boldsymbol{y}} p\left(\boldsymbol{x} \mid \boldsymbol{y}\right)  \mathrm{d} \boldsymbol{x} \\
= & \nabla_{\boldsymbol{y}} \int p\left(\boldsymbol{x} \mid \boldsymbol{y}\right) \mathrm{d} \boldsymbol{x} \\
= & \nabla_{\boldsymbol{y}} 1 =0
\end{align*}
Thus, \cref{eq:score-function} is proved.
\end{proof}

\subsection{A useful Lemma}

\begin{lemma}
\label{lemma:1}
We state the probability density transform equation as follows. Suppose $\boldsymbol{x} \sim \mathbf{x}$ and $\boldsymbol{y} \sim \mathbf{y}$ and $\boldsymbol{y} = f(\boldsymbol{x})$. Assume $f$ is invertible and its inverse function is $g$. Then, we have
\begin{equation*}
    p_{\mathbf{y}}\left( \boldsymbol{y}\right) = \left|\frac{\partial g}{\partial \boldsymbol{y}}\right| p_{\mathbf{x}}\left( g(\boldsymbol{y})\right).
\end{equation*}
\end{lemma}

\subsection{The proof of \cref{eq:gamma-log} in Sec. 3.1.3}
\begin{equation}
\begin{split}
    \boldsymbol{f} \left(\boldsymbol{x}, \boldsymbol{y}\right) = \nabla_{\boldsymbol{y}} \log p\left(\boldsymbol{y} \mid \boldsymbol{x}\right) = \frac{\alpha - 1}{\boldsymbol{y}} - \frac{\alpha}{\boldsymbol{x}}.
\end{split}
\tag{8}
\end{equation}
\begin{proof}
According to \cref{eq:gamma-prob}, we have that:
\begin{align*}
& \nabla_{\boldsymbol{y}} \log p\left(\boldsymbol{y} \mid \boldsymbol{x}\right) \\
= & \nabla_{\boldsymbol{y}} \log \prod_{i=1}^{d} \frac{\alpha^\alpha}{\Gamma\left( \alpha \right)} \left(\frac{y_i}{x_i}\right)^{\alpha - 1} \exp \left\{- \frac{\alpha y_i}{x_i}\right\} \cdot \frac{1}{x_i} \\
=& \nabla_{\boldsymbol{y}} \sum_{i=1}^{d} \log \frac{\alpha^\alpha}{\Gamma\left( \alpha \right)} \left(\frac{y_i}{x_i}\right)^{\alpha - 1} \exp \left\{- \frac{\alpha y_i}{x_i}\right\} \cdot \frac{1}{x_i} \\
=& \sum_{i=1}^{d} \nabla_{\boldsymbol{y}} \log \frac{\alpha^\alpha}{\Gamma\left( \alpha \right)} \left(\frac{y_i}{x_i}\right)^{\alpha - 1} \exp \left\{- \frac{\alpha y_i}{x_i}\right\} \cdot \frac{1}{x_i} \\
=& \sum_{i=1}^{d} \nabla_{\boldsymbol{y}} \left( (\alpha - 1) \log y_i - \frac{\alpha y_i}{x_i} \right) \\
=& \frac{\alpha - 1}{\boldsymbol{y}} - \frac{\alpha}{\boldsymbol{x}}.
\end{align*}
\end{proof}

\subsection{The proof of \cref{eq:gamma-solution} in Sec. 3.1.3}

\begin{equation}
    \hat{\boldsymbol{x}} = \frac{\alpha \boldsymbol{y}}{\alpha - 1 - \boldsymbol{y} \odot \boldsymbol{s}\left( \boldsymbol{y}\right)}. \tag{9}
\end{equation}

\begin{proof}
\begin{align*}
& \boldsymbol{s}\left( \boldsymbol{y}\right) = \frac{\alpha - 1}{\boldsymbol{y}} - \frac{\alpha}{\boldsymbol{x}} \\
\Longleftrightarrow & \boldsymbol{y} \odot \boldsymbol{s}\left( \boldsymbol{y}\right) = \alpha - 1 - \frac{\alpha \boldsymbol{y}}{\boldsymbol{x}} \\
\Longleftrightarrow & \frac{\alpha \boldsymbol{y}}{\boldsymbol{x}}   = \alpha - 1 - \boldsymbol{y} \odot \boldsymbol{s}\left( \boldsymbol{y}\right) \\
\Longleftrightarrow & {\boldsymbol{x}} = \frac{\alpha \boldsymbol{y}}{\alpha - 1 - \boldsymbol{y} \odot \boldsymbol{s}\left( \boldsymbol{y}\right)}.
\end{align*}
\end{proof}

\subsection{The proof of \cref{eq:poisson-log} in Sec. 3.1.3}

\begin{equation}
\begin{split}
    \boldsymbol{f} \left(\boldsymbol{x}, \boldsymbol{y}\right) &= \nabla_{\boldsymbol{y}} \log \mathrm{Pr} \left(\boldsymbol{y} \mid \boldsymbol{x}\right) \\
    &= \lambda \log \left( \lambda \boldsymbol{x} \right) - \lambda \log \left( \lambda \boldsymbol{y} + \frac{1}{2} \right).
\end{split}
\tag{12}
\end{equation}

\begin{proof}
According to \cref{eq:poisson-prob}, we have that
\begin{align*}
& \nabla_{\boldsymbol{y}} \log \mathrm{Pr} \left(\boldsymbol{y} \mid \boldsymbol{x}\right) \\
= & \nabla_{\boldsymbol{y}} \log \prod_{i=1}^{d} \frac{\left( \lambda x_{i} \right)^{\lambda_{i} y_{i}}}{\left( \lambda y_{i} \right) !} e^{- \lambda x_{i}} \\
= & \sum_{i=1}^{d} \nabla_{\boldsymbol{y}} \log  \frac{\left( \lambda x_{i} \right)^{\lambda_{i} y_{i}}}{\left( \lambda y_{i} \right) !} e^{- \lambda x_{i}} \\
= & \sum_{i=1}^{d} \nabla_{\boldsymbol{y}} \left( \lambda_{i} y_{i} \log  \lambda x_{i} - \log \left( \lambda y_{i} \right) ! \right) \\
= & \lambda \log \left( \lambda \boldsymbol{x} \right) - \lambda \log \left( \lambda \boldsymbol{y} + \frac{1}{2} \right) \\
\end{align*}
Here, we set $\nabla_{y_i} \log \left( \lambda y_{i} \right) ! = \lambda \log \left( \lambda {y}_{i} + \frac{1}{2} \right)$.
\end{proof}

\subsection{The proof of \cref{eq:poisson-solution} in Sec. 3.1.3}

\begin{equation}
    \hat{\boldsymbol{x}} = \left( \boldsymbol{y} + \frac{1}{2\lambda} \right) \odot \exp \left\{\frac{\boldsymbol{s}\left( \boldsymbol{y}\right)}{\lambda } \right\}. \tag{13}
\end{equation}

\begin{proof}
\begin{align*}
& \boldsymbol{s}\left( \boldsymbol{y}\right) = \lambda \log \left( \lambda \boldsymbol{x} \right) - \lambda \log \left( \lambda \boldsymbol{y} + \frac{1}{2} \right) \\
\Longleftrightarrow & \frac{\boldsymbol{s}\left( \boldsymbol{y}\right)}{\lambda} =  \log \left( \lambda\boldsymbol{x} \right) - \log \left( \lambda \boldsymbol{y} + \frac{1}{2} \right) \\
\Longleftrightarrow  & \log \left( \lambda \boldsymbol{x} \right) =  \frac{\boldsymbol{s}\left( \boldsymbol{y}\right)}{\lambda} + \log \left( \lambda \boldsymbol{y} + \frac{1}{2} \right) \\
\Longleftrightarrow  & \lambda \boldsymbol{x} =  \left( \lambda \boldsymbol{y} + \frac{1}{2} \right) \exp \left\{ \frac{\boldsymbol{s}\left( \boldsymbol{y}\right)}{\lambda} \right\} \\
\Longleftrightarrow  & \boldsymbol{x} =  \left( \boldsymbol{y} + \frac{1}{2\lambda } \right) \exp \left\{ \frac{\boldsymbol{s}\left( \boldsymbol{y}\right)}{\lambda} \right\} 
\end{align*}

\end{proof}

\subsection{The proof of \cref{eq:rayleigh-log} in Sec. 3.1.3}

\begin{equation}
\begin{split}
    \boldsymbol{f} \left(\boldsymbol{x}, \boldsymbol{y}\right) &= \nabla_{\boldsymbol{y}} \log p \left(\boldsymbol{y} \mid \boldsymbol{x}\right) = \frac{1}{\boldsymbol{y} - \boldsymbol{x}} - \frac{\boldsymbol{y} - \boldsymbol{x}}{\sigma^2 \boldsymbol{x}^2}.
\end{split}
\tag{16}
\end{equation}

\begin{proof}
According to \cref{eq:rayleigh-prob}, we have that
\begin{align*}
& \nabla_{\boldsymbol{y}} \log p \left(\boldsymbol{y} \mid \boldsymbol{x}\right) \\
= & \nabla_{\boldsymbol{y}} \log \prod_{i=1}^{d} \frac{1}{x_{i}} \frac{y_{i} - x_{i}}{x_{i} \sigma^2} \exp \left\{ - \frac{(y_i - x_i)^2}{2 x_i^2 \sigma^2}\right\} \\
= & \sum_{i=1}^{d} \nabla_{\boldsymbol{y}} \log  \frac{1}{x_{i}} \frac{y_{i} - x_{i}}{x_{i} \sigma^2} \exp \left\{ - \frac{(y_i - x_i)^2}{2 x_i^2 \sigma^2}\right\} \\
= & \sum_{i=1}^{d} \nabla_{\boldsymbol{y}} \left( \log \left(y_{i} - x_{i}\right) - \frac{(y_i - x_i)^2}{2 x_i^2 \sigma^2} \right) \\
= & \frac{1}{\boldsymbol{y} - \boldsymbol{x}} - \frac{\boldsymbol{y} - \boldsymbol{x}}{\sigma^2 \boldsymbol{x}^2}.
\end{align*}
\end{proof}

\subsection{The proof of the Solving Method in Algorithm\cref{alg:rayleigh-iterative}}
\begin{proof}
Our target equation is
\begin{equation*}
\boldsymbol{s}\left( \boldsymbol{y}\right) = \frac{1}{\boldsymbol{y} - \boldsymbol{x}} - \frac{\boldsymbol{y} - \boldsymbol{x}}{\sigma^2 \boldsymbol{x}^2}.    
\end{equation*}
For simplicity, we do not use bold font. Let $ t = \frac{y - x}{x}$ and assume $t > 0$ because $x$ should be smaller than $y$ according to the Rayleigh distribution. We denote $\boldsymbol{s}\left( \boldsymbol{y}\right)$ as $s$. Fixing $x$, then
\begin{align*}
& {s} = \frac{1}{{y} - {x}} - \frac{{y} - {x}}{\sigma^2 {x}^2} \\
\Longleftrightarrow &  sx = \frac{x}{y - x} - \frac{y - x}{\sigma^2 x} \\
\Longleftrightarrow &  sx = \frac{1}{t} - \frac{t}{\sigma^2}  \\
\Longleftrightarrow & t^2 + \sigma^2 s x t - \sigma^2 = 0
\end{align*}
Since $t > 0$, we have 
\begin{equation*}
t = \frac{- \sigma^2 s x + \sqrt{\sigma^4 s^2x^2 + 4 \sigma^2}}{2}
\end{equation*}
After solving $t$, we compute $x = \frac{y}{t + 1}$. Therefore, the iterative process contains two steps:
\begin{itemize}
    \item $t = \frac{- \sigma^2 s x + \sqrt{\sigma^4 s^2x^2 + 4 \sigma^2}}{2}$
    \item $x = \frac{y}{t + 1}$
\end{itemize}
\end{proof}

\subsection{The Proof of \cref{eq:multi-conv} in Sec. 3.1.3}

\begin{equation}
\begin{split}
    \nabla_{\boldsymbol{y}} \log p_{\mathbf{y}} \left(\boldsymbol{y} \mid \boldsymbol{x}\right) = \boldsymbol{A}^{-1, \top} \nabla_{\boldsymbol{z}} \log p_{\mathbf{z}} \left( \boldsymbol{A}^{-1} \boldsymbol{y} \mid \boldsymbol{x} \right).
\end{split}
\tag{18}
\end{equation}

\begin{proof}
According to \cref{lemma:1}, we have $\boldsymbol{y} = f(\boldsymbol{z}) = \boldsymbol{A}^{-1}\boldsymbol{z}$, then $g(\boldsymbol{y}) = \boldsymbol{A}\boldsymbol{y}$. Thus,
\begin{equation*}
    p_{\mathbf{y}} \left(\boldsymbol{y} \mid \boldsymbol{x}\right) = \left| \boldsymbol{A}^{-1} \right| \nabla_{\boldsymbol{z}} \log p_{\mathbf{z}} \left( \boldsymbol{A}^{-1} \boldsymbol{y} \mid \boldsymbol{x} \right).
\end{equation*}
Obviously, \cref{eq:multi-conv} is proved.
\end{proof}

\subsection{The Proof of \cref{eq:mixture-taylor} in Sec 3.1.4}

\begin{equation}
\begin{split}
    p_{\mathbf{y}} \left( \boldsymbol{y} \mid \boldsymbol{x} \right)
    \approx  p_{\mathbf{z}} \left( \bar{\boldsymbol{z}} \mid \mathbf{x} \right) + \nabla_{\boldsymbol{z}} p_{\mathbf{z}} \left( \bar{\boldsymbol{z}} \mid \boldsymbol{x} \right)^T \left( \boldsymbol{y} - \bar{\boldsymbol{z}} \right).
\end{split}
\tag{21}
\end{equation}

\begin{proof}
\begin{equation*}
\begin{split}
    & p_{\mathbf{y}} \left( \boldsymbol{y} \mid \boldsymbol{x} \right) \\
    \approx &  \int_{\boldsymbol{z} \approx \boldsymbol{y}} p_{\mathbf{y}} \left( \boldsymbol{y} \mid \boldsymbol{z} \right) p_{\mathbf{z}} \left( \boldsymbol{z} \mid \boldsymbol{x} \right) \mathrm{d} \boldsymbol{z} \\
    \approx & \int_{\boldsymbol{z} \approx \boldsymbol{y}} p_{\mathbf{y}} \left( \boldsymbol{y} \mid \boldsymbol{z} \right) \left( p_{\mathbf{z}} \left( \bar{\boldsymbol{z}} \mid \boldsymbol{x} \right) + \nabla_{\boldsymbol{z}} p_{\mathbf{z}} \left( \bar{\boldsymbol{z}} \mid \boldsymbol{x} \right)^T \left(\mathbf{z} - \bar{\boldsymbol{z}} \right) \right)  \mathrm{d} \boldsymbol{z} \\
    = & p_{\mathbf{z}} \left( \bar{\boldsymbol{z}} \mid \boldsymbol{x} \right) \int_{\boldsymbol{z} \approx \boldsymbol{y}} p_{\mathbf{y}} \left( \boldsymbol{y} \mid \boldsymbol{z} \right)\mathrm{d} \boldsymbol{z} \\
    & \quad \quad + \nabla_{\boldsymbol{z}} p_{\mathbf{z}} \left( \bar{\boldsymbol{z}} \mid \boldsymbol{x} \right)^T \int_{\boldsymbol{z} \approx \boldsymbol{y}} p_{\mathbf{y}} \left( \boldsymbol{y} \mid \boldsymbol{z} \right) \left(\boldsymbol{z} - \bar{\boldsymbol{z}} \right)  \mathrm{d} \boldsymbol{z} \\
    \approx & p_{\mathbf{z}} \left( \bar{\boldsymbol{z}} \mid \mathbf{x} \right) + \nabla_{\boldsymbol{z}} p_{\mathbf{z}} \left( \bar{\boldsymbol{z}} \mid \boldsymbol{x} \right)^T \left( \boldsymbol{y} - \bar{\boldsymbol{z}} \right).
\end{split}
\end{equation*}    
\end{proof}

\subsection{The Proof of \cref{eq:mixture-score} in Sec 3.1.4}

\begin{equation}
    \nabla_{\boldsymbol{y}} \log p_{\mathbf{y}} \left( \boldsymbol{y} \mid \boldsymbol{x} \right) \approx  \nabla_{\boldsymbol{z}} \log p_{\mathbf{z}} \left( \bar{\boldsymbol{z}} \mid \boldsymbol{x} \right). \tag{22}
\end{equation}

\begin{proof}
\begin{align*}
& \nabla_{\boldsymbol{y}} \log p_{\mathbf{y}} \left( \boldsymbol{y} \mid \boldsymbol{x} \right) \\
=& \nabla_{\boldsymbol{y}} \log \left( p_{\mathbf{z}} \left( \bar{\boldsymbol{z}} \mid \boldsymbol{x} \right)\left( 1 + \frac{\nabla_{\boldsymbol{z}} p_{\mathbf{z}} \left( \bar{\boldsymbol{z}} \mid \boldsymbol{x} \right)^T \left( \boldsymbol{y} - \bar{\boldsymbol{z}} \right)}{p_{\mathbf{z}} \left( \bar{\boldsymbol{z}} \mid \boldsymbol{x} \right)} \right) \right) \\
\approx &  \nabla_{\boldsymbol{y}} \log p_{\mathbf{z}} \left( \bar{\boldsymbol{z}} \mid \boldsymbol{x} \right) + \nabla_{\boldsymbol{y}} \frac{\nabla_{\boldsymbol{z}} p_{\mathbf{z}} \left( \bar{\boldsymbol{z}} \mid \boldsymbol{x} \right)^T \left( \boldsymbol{y} - \bar{\boldsymbol{z}} \right)}{p_{\mathbf{z}} \left( \bar{\boldsymbol{z}} \mid \boldsymbol{x} \right)} \\
= &\frac{\nabla_{\boldsymbol{z}} p_{\mathbf{z}} \left( \bar{\boldsymbol{z}} \mid \boldsymbol{x} \right)}{p_{\mathbf{z}} \left( \bar{\boldsymbol{z}} \mid \boldsymbol{x} \right)} = \nabla_{\boldsymbol{z}} \log p_{\mathbf{z}} \left( \bar{\boldsymbol{z}} \mid \boldsymbol{x} \right)
\end{align*}    
\end{proof}

\subsection{The Proof of \cref{eq:tweedie-score} in Sec. 3.2}

\begin{equation}
\begin{split}
\boldsymbol{s} \left( \boldsymbol{y} \right) = \nabla_{{\boldsymbol{y}}} \log b \left({\boldsymbol{y}}\right) + T^{\prime}\left( \boldsymbol{y} \right)^{\top} \mathbb{E} \left[ H(\boldsymbol{x}) \mid {\boldsymbol{y}}\right].
\end{split}
\tag{28}
\end{equation}

\begin{proof}
\begin{equation*}
\begin{split}
\boldsymbol{s} \left( \boldsymbol{y} \right) &= \nabla_{\boldsymbol{y}} \log p\left(\boldsymbol{y}\right) \\
&= \int p\left(\boldsymbol{x} \mid \boldsymbol{y}\right) \nabla_{\boldsymbol{y}} \log p\left(\boldsymbol{y} \mid \boldsymbol{x}\right) \mathrm{d} \boldsymbol{x} \\
& = \int p\left(\boldsymbol{x} \mid \boldsymbol{y}\right) \left( \nabla_{{\boldsymbol{y}}} \log b \left({\boldsymbol{y}}\right) + \nabla_{\boldsymbol{y}} H(\boldsymbol{x})^{\top} T\left( \boldsymbol{y} \right) \right) \mathrm{d} \boldsymbol{x} \\
& = \nabla_{{\boldsymbol{y}}} \log b \left({\boldsymbol{y}}\right) + T^{\prime}\left( \boldsymbol{y} \right)^{\top} \int p\left(\boldsymbol{x} \mid \boldsymbol{y}\right) H(\boldsymbol{x}) \mathrm{d} \boldsymbol{x} \\
&= \nabla_{{\boldsymbol{y}}} \log b \left({\boldsymbol{y}}\right) + T^{\prime}\left( \boldsymbol{y} \right)^{\top} \mathbb{E} \left[ H(\boldsymbol{x}) \mid {\boldsymbol{y}}\right].
\end{split}
\end{equation*}
\end{proof}

\section{Experiment}
\label{sec:app-experiment}

When training score function, for $\sigma_a$ in Eq. (29), we set initial value as $0.05$ and final value as $1 \times 10^{-6}$. We reduce $\sigma_a$ linearly every 50 training steps and keep it as $1 \times 10^{-6}$ for the final 50 steps. Another important point about the training for non-Gaussian noise model (from No.$5$ to No.$10$), we add a slight Gaussian noise to noisy images such that the score function estimation is stable and remove the additive Gaussian noise when inference as we do in mixture noise models. Here, we set the $\sigma$ of Gaussian noise as $5$.

For Neighbor2Neighbor, We use the code in \hyperlink{https://github.com/TaoHuang2018/Neighbor2Neighbor}{https://git-hub.com/TaoHuang2018/Neighbor2Neighbor} and keep the default hyper-parameters setting.

For Noisier2Noise, we use the code in \hyperlink{https://github.com/melobron/Noisier2Noise}{https://git-hub.com/melobron/Noisier2Noise}. We set $\alpha = 1$ and compute the average of $50$ denoised results.

\end{document}